\documentclass[11pt,a4paper]{article}
\usepackage[latin9]{inputenc}
\usepackage{booktabs}
\usepackage{calc}
\usepackage{units}
\usepackage{amstext}
\usepackage{graphicx}

\makeatletter

\pdfpageheight\paperheight
\pdfpagewidth\paperwidth

\providecommand{\tabularnewline}{\\}

\@ifundefined{date}{}{\date{}}
\usepackage{acl2018}
\usepackage{times}
\usepackage{latexsym}

\usepackage{url}

\aclfinalcopy % Uncomment this line for the final submission
\title{Towards a Neural Network Approach to Abstractive \\ Multi-Document Summarization}

\author{Jianmin Zhang \and Jiwei Tan \and Xiaojun Wan \\
  Institute of Computer Science and Technology, Peking University \\
  The MOE Key Laboratory of Computational Linguistics, Peking University \\
  {\tt \{zhangjianmin2015,tanjiwei,wanxiaojun\}@pku.edu.cn} \\}

\makeatother

\begin{document}
\maketitle
\begin{abstract}
Till now, neural abstractive summarization methods have achieved great
success for single document summarization (SDS). However, due to the
lack of large scale multi-document summaries, such methods can be
hardly applied to multi-document summarization (MDS). In this paper,
we investigate neural abstractive methods for MDS by adapting a state-of-the-art
neural abstractive summarization model for SDS. We propose an approach
to extend the neural abstractive model trained on large scale SDS
data to the MDS task. Our approach only makes use of a small number
of multi-document summaries for fine tuning. Experimental results
on two benchmark DUC datasets demonstrate that our approach can outperform
a variety of baseline neural models.
\end{abstract}

\section{Introduction}

Document summarization is a task of automatically producing a summary
for given documents. Different from Single Document Summarization
(SDS) which generates a summary for each given document, Multi-Document
Summarization (MDS) aims to generate a summary for a set of topic-related
documents. Previous approaches to document summarization can be generally
categorized to extractive methods and abstractive methods. Extractive
methods produce a summary by extracting and merging sentences from
the original document(s), while abstractive methods generate a summary
using arbitrary words and expressions based on understanding the document(s).
Due to the difficulty of natural language understanding and generation,
previous research on document summarization is more focused on extractive
methods \cite{yao2017recent}. However, extractive methods suffer
from the inherent drawbacks of discourse incoherence and long, redundant
sentences, which hampers its application in reality \cite{tan2017abstractive}.
Recently, with the success of sequence-to-sequence (seq2seq) models
in natural language generation tasks including machine translation
\cite{bahdanau2014neural} and dialog systems \cite{mou2016sequence},
abstractive summarization methods has received increasing attention.
With the resource of large-scale corpus of human summaries, it is
able to train an abstractive summarization model in an end-to-end
framework. Neural abstractive summarization models \cite{DBLP:conf/acl/SeeLM17,tan2017abstractive}
have surpass the performance of extractive methods on single document
summarization task with abundant training data.

Unfortunately, the extension of seq2seq models to MDS is not straightforward.
Neural abstractive summarization models are usually trained on about
hundreds of thousands of gold summaries, but there are usually very
few human summaries available for the MDS task. More specifically,
in the news domain, there is only a few hundred multi-document summaries
provided by DUC and TAC conferences in total, which are largely insufficient
for training neural abstractive models. Apart from insufficient training
data, neural models for abstractive MDS also face the challenge of
much more input content, and the study is still in the primary stage.

In this study, we investigate applying seq2seq models to the MDS task.
We attempt various ways of extending neural abstractive summarization
models pre-trained on the SDS data to the MDS task, and reveal that
neural abstractive summarization models do not transfer well on a
different dataset. Then we study the factors which affect the transfer
performance, and propose methods to adapt the pre-trained model to
the MDS task. We also study leveraging the few MDS training data to
further improve the pre-trained model. We conduct experiment on the
benchmark DUC datasets, and experiment results demonstrate our approach
is able to achieve considerable improvement over a variety of neural
baselines.

The contributions of this study are summarized as follows:
\begin{itemize}
\item To the best of our knowledge, our work is one of the very few pioneering
works to investigate adapting neural abstractive summarization models
of single document summarization to the task of multi-document summarization.
\item We propose a novel approach to adapt the neural model trained on the
SDS data to the MDS task, and leverage the few MDS training data to
further improve the pre-trained model.
\item Evaluation results demonstrate the efficacy of our proposed approach,
which outperforms a variety of neural baselines.
\end{itemize}
We organize the paper as follows. In Section 2 we introduce related
work. In Section 3 we describe the previous neural abstractive summarization
model. Then we introduce our proposed approach in Section 4. Experiment
results and discussion are presented in Section 5. Finally, we conclude
this paper in Section 6.

\section{Related Work}

\subsection{Extractive Summarization Methods}

The study of MDS is pioneered by \cite{DBLP:conf/sigir/McKeownR95},
and early notable works also include \cite{DBLP:conf/aaai/McKeownKHBE99,radev2000centroid}.
Extractive summarization systems that compose a summary from a number
of important sentences from the source documents are by far the most
popular solution for MDS \cite{DBLP:conf/acl/AvineshM17}. Redundancy
is one of the biggest problems for extractive methods \cite{DBLP:journals/air/GambhirG17},
and the Maximal Marginal Relevance (MRR) \cite{DBLP:conf/sigir/CarbonellG98}
is a well-known algorithm for reducing redundancy. In the past years
various models under extractive framework have been proposed \cite{DBLP:conf/skg/TaoZLG08,DBLP:conf/sigir/WanY08,DBLP:journals/tkdd/WangZLCG11,DBLP:conf/sigir/TanWX15}.
One important architecture is to model MDS as a budgeted maximum coverage
problem, including the prior approach \cite{DBLP:conf/ecir/McDonald07}
and improved models \cite{DBLP:conf/emnlp/WoodsendL12,DBLP:conf/acl/LiQL13,DBLP:conf/emnlp/BoudinMF15}.
There are still recent studies under traditional extractive framework
\cite{DBLP:conf/acl/PeyrardE17,DBLP:conf/acl/AvineshM17}.

\subsection{Abstractive Summarization Methods}

Abstractive summarization methods aim at generating the summary based
on understanding the original documents. Sequence-to-sequence models
with attention mechanism have been applied to the abstractive summarization
task. Success attempts are on sentence summarization \cite{rush2015neural,DBLP:conf/naacl/ChopraAR16,nallapati2016abstractive}
or single document summarization \cite{tan2017abstractive,DBLP:conf/acl/SeeLM17,DBLP:journals/corr/PaulusXS17},
which have abundant gold summaries to train an end-to-end system.

Until very recent, there occurs attempt for abstractive multi-document
summarization under the seq2seq framework. The lack of enough train
examples is the major obstacle to this end. To address this, \newcite{DBLP:journals/corr/abs-1801-10198}
study the task of generating English Wikipedia under a viewpoint of
multi-document summarization. They construct a large corpus with reference
summaries, so that end-to-end training of a seq2seq is capable. Their
study reveals that seq2seq model works when there are abundant training
data for MDS. Very recently \newcite{DBLP:journals/corr/abs-1801-07704}
try to apply pre-trained abstractive summarization model of SDS to
the query-focused summarization task. They  sort the input
documents and then iteratively apply the SDS model to summarize each
single document until the length limit is reached. Their major concern
is incorporating query information into the abstractive model or using
the query to filter the original documents, which is different from
our work focusing on generic multi-document summarization. Moreover,
the intuitive idea of using the SDS model for summarizing each single
document in the multi-document set is adopted in the baseline models
for comparison as well.

\section{Preliminaries}

\begin{figure}
\includegraphics[width=0.95\columnwidth]{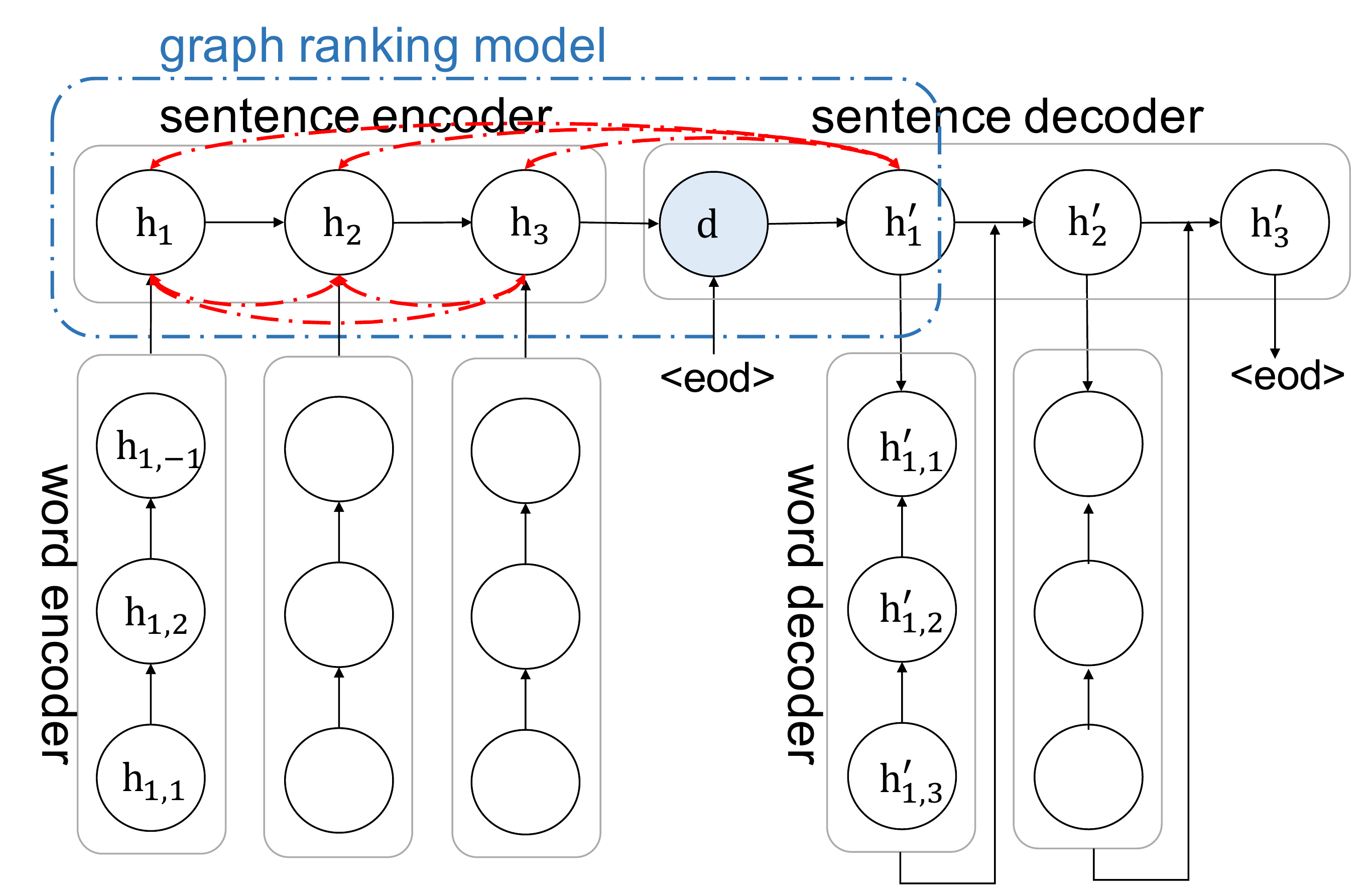}

\caption{\label{fig:SinABS-model.}SinABS model. The figure is borrowed from
\newcite{tan2017abstractive}. }
\end{figure}

In this work we investigate abstractive MDS approach based on the
state-of-the-art neural abstractive model in \newcite{tan2017abstractive}.
Compared with another neural abstractive model in \newcite{DBLP:conf/acl/SeeLM17},
\newcite{tan2017abstractive} adopt a hierarchical encoder-decoder
framework which we found is more scalable to more and longer input
documents. The model is named \textbf{SinABS} in this paper. SinABS
uses a hierarchical encoder-decoder framework like \newcite{li2015hierarchical},
where a PageRank \cite{page1999pagerank} based attention mechanism
is proposed to identify salient sentences in the original documents.
The SinABS model is illustrated in Figure~\ref{fig:SinABS-model.}.We
introduce the SinABS model following \newcite{tan2017abstractive}.

\subsection{Encoder}

The target of the encoder is to encode the input documents into vector
representations. SinABS adopts a hierarchical encoder framework, where
a word encoder $enc_{\mathrm{word}}$ is used for encoding a sentence
into the sentence representation from its words, as $\mathbf{h}_{i,k}=enc_{\mathrm{word}}(\mathbf{h}_{i,k-1},\mathbf{e}_{i,k})$,
where $\mathbf{h}_{i,k}$ represents the hidden state when LSTM receives
word $\mathbf{e}_{i,k}$. Then a sentence encoder $enc_{\mathrm{sent}}$
is used for encoding an input document into the document representation
from its sentences, as $\mathbf{h}_{i}=enc_{\mathrm{sent}}(\mathbf{h}_{i-1},\mathbf{x}_{i})$,
where $\mathbf{x}_{i}=\mathbf{h}_{i,-1}$ is the last hidden state
when word encoder receives the whole sentence $i$. The input to the
word encoder is the word sequence of a sentence, appended with an
``\textless{}eos\textgreater{}'' token indicating the end of
a sentence. The last hidden state after the word encoder receives
``\textless{}eos\textgreater{}'' is used as the embedding representation
of the sentence. A sentence encoder is used to sequentially receive
the embeddings of the sentences. A pseudo sentence of an ``\textless{}eod\textgreater{}''
token is appended at the end of the document to indicate the end of
the whole document. The hidden state after the sentence encoder receives
``\textless{}eod\textgreater{}'' is treated as the representation
of the input document, denoted as $\mathbf{c}$. Long Short-Term Memory
(LSTM) \cite{hochreiter1997long} is used as the word encoder $enc_{\mathrm{word}}$
and also the sentence encoder $enc_{\mathrm{sent}}$.

\subsection{Decoder}

Similar to the hierarchical encoder, The sentence decoder $dec_{\mathrm{sent}}$
receives the document representation $\mathbf{d}$ as the initial
state $\mathbf{h}_{0}^{'}=\mathbf{d}$, and predicts the sentence
representations sequentially, by $\mathbf{h}_{j}^{'}=dec_{\mathrm{sent}}(\mathbf{h}_{j-1}^{'},\mathbf{x}_{j-1}^{'})$,
where $\mathbf{x}_{j-1}^{'}$ is the encoded representation of the
previously generated sentence $s_{j-1}^{'}$. The word decoder $dec_{\mathrm{word}}$
receives a sentence representation $\mathbf{h}_{j}^{'}$ as the initial
state $\mathbf{h}_{j,0}^{'}=\mathbf{h}_{j}^{'}$, and predicts the
word representations sequentially, by $\mathbf{h}_{j,k}^{'}=dec_{\mathrm{word}}(\mathbf{h}_{j,k-1}^{'},\mathbf{e}_{j,k-1})$,
where $\mathbf{e}_{j,k-1}$ is the embedding of the previously generated
word. The predicted word representations are first concatenated with
the context vector $\mathbf{c}_{j}$, and then mapped to vectors of
the vocabulary size dimension by a projection layer, and finally normalized
by a softmax layer as the probability distribution of generating the
words in the vocabulary. A word decoder stops when it generates the
``\textless{}eos\textgreater{}'' token and similarly the sentence
decoder stops when it generates the ``\textless{}eod\textgreater{}''
token.

\subsection{Attention Mechanism}

The attention mechanism used in SinABS sets a different context vector
$\mathbf{c}_{j}$ when generating the words of sentence $j$, by $\mathbf{c}_{j}=\sum_{i}\alpha_{i}^{j}\mathbf{h}_{i}$.
The graph-based attention mechanism in \newcite{tan2017abstractive}
adopts the topic-sensitive PageRank algorithm to compute the attention
weights, by

\begin{equation}
\mathbf{f}=(1-\lambda)(I-\lambda WD^{-1})^{-1}\mathbf{y}\label{eq:close-form}
\end{equation}
where $\mathbf{f}=[f_{1},\ldots,f_{n}]\in\mathcal{R}^{n}$ denotes
the rank scores of the $n$ original sentences. $D$ is a diagonal
matrix with its $\left(i,i\right)$-element equal to the sum of the
$i$-th column of $W$. $W(i,j)=\mathbf{h}_{i}^{T}P\mathbf{h}_{j}$
where $P$ is a parameter matrix to be learned. $\lambda$ is a damping
factor and set to 0.9. $\mathbf{y}\in\mathcal{R}^{n}$ is a one hot
vector and only $y_{0}=1$. The ranked scores are then integrated
with a distraction mechanism, and finally computed as:

\begin{equation}
\alpha_{i}^{j}=\frac{\max(f_{i}^{j}-f_{i}^{j-1},0)}{\sum_{l}\left(\max(f_{l}^{j}-f_{l}^{j-1},0)\right)}\label{eq:diffatt}
\end{equation}

\section{Our Approach}

\subsection{Overview}

\begin{figure}
\includegraphics[width=1\columnwidth]{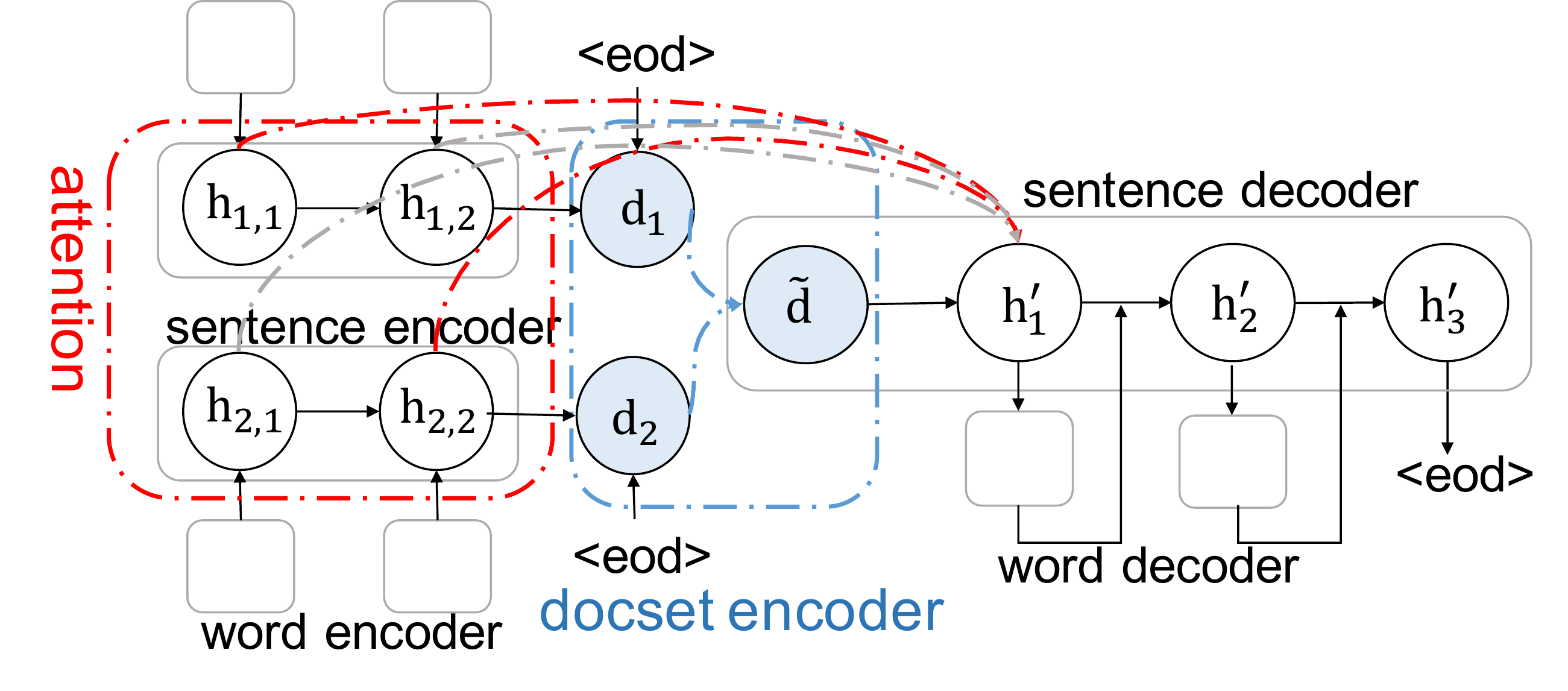}

\caption{\label{fig:Framework-our}Framework of our model. The difference from
Figure~\ref{fig:SinABS-model.} is the docset encoder and the concentrated
attention mechanism.}

\end{figure}

In this section we introduce our approach. Our abstractive MDS model
is the extension of the single document summarization model SinABS.
It is an encoder-decoder framework, which takes all the documents
of a document set as input, then encodes the documents into a document
set representation, and further generates the summary with a decoder.
To adapt SinABS to the MDS task, our model is different from SinABS
in the encoder model and the attention mechanism, and it will also
be tuned on the MDS dataset to adapt to the MDS task. The framework
of our model is illustrated in Figure~\ref{fig:Framework-our}.

\subsection{Multi-Document Encoder}

The major difference of MDS is that we need to generate a summary
for multiple input documents. So our system needs to deal with the
multiple input documents although SinABS is trained to generate a
summary for one document. Considering that the decoder generates the
summary from the representation vector encoded by the encoder, we
can generate a summary for a document set if the document set is encoded
to a representation vector containing its key information. In our approach, we achieve this by adding a document set encoder, to encode
a set of document representation vectors into a document set representation.
Thus the hierarchical encoder structure becomes three levels.

The document set encoder $enc_{\textrm{docset}}$ takes document vectors
$\{\mathbf{d}_{m}\}$, $m\in[1,M]$ where $M$ is the number of documents
in a document set as input, produces a new document set vector $\tilde{\mathbf{d}}$,
and then $\tilde{\mathbf{d}}$ is provided to the decoder to generate
the summary for the document set. The decoder will be a two-level
hierarchical framework similar to that in \newcite{tan2017abstractive}.
Since there is no order and dependency relationship between different
documents in a document set, it is not reasonable to use LSTM as the
document set encoder. Instead, we define the document set encoder
as:

\begin{equation}
\tilde{\mathbf{d}}=enc_{\textrm{docset}}\left(\{\mathbf{d}_{m}\}\right)=\sum_{m}w_{m}\mathbf{d}_{m}
\end{equation}
where $\mathbf{w}=[w_{1},\ldots,w_{m}]\in\mathcal{R}^{m}$ is a weight
vector to merge the document vectors into a document set representation.
The weight vector $\mathbf{w}$ can be a fixed one as $\mathbf{w}=[\nicefrac{1}{m},\ldots,\nicefrac{1}{m}]$,
but in our system we hope to assign different $w_{m}$ to different
$\mathbf{d}_{m}$, since different documents may contribute differently
to the overall summary. However, it is unreasonable to treat $\mathbf{w}$
as a parameter vector and learn it directly, because the weight $w_{m}$
for $\mathbf{d}_{m}$ should be based on $\mathbf{d}_{m}$. The position
of a document should not affect its weight since there is no order
in a document set.

In our system the weight for a document is decided based on the document
itself, and its contribution to the representation of the overall
document set. Therefore, we define:

\begin{equation}
w_{m}=\frac{\mathbf{q}^{T}[\mathbf{d}_{m};\mathbf{d}_{\Sigma}]}{\sum_{m}\mathbf{q}^{T}[\mathbf{d}_{m};\mathbf{d}_{\Sigma}]}\label{eq:wm}
\end{equation}
where $\mathbf{d}_{\Sigma}=\sum_{m}\mathbf{d}_{m}$ and $[\mathbf{d}_{m};\mathbf{d}_{\Sigma}]$
is the concatenation of $\mathbf{d}_{m}$ and $\mathbf{d}_{\Sigma}$.
The intuitive explanation of Eq.~\ref{eq:wm} is that the weight of
$\mathbf{d}_{m}$ is decided by its relationship (modeled by parameterized
dot product) with the representation of the whole document set $\mathbf{d}_{\Sigma}$.
$\mathbf{q}$ is the parameter to be learned, whose dimension is twice
the dimension of $\mathbf{d}_{m}$ or $\mathbf{d}_{\Sigma}$.

\subsection{Attention}

The decoder receives the document set vector $\tilde{\mathbf{d}}$
as initial state and generates the output summary from the document
set representation. The difference of the decoder to SinABS is that
when computing the attention distribution now it should be computed
on all the sentences in a document set. Not only the amount of original
sentences becomes larger, but also the original sentences come from
different documents. Nevertheless, we believe the topic-sensitive
PageRank attention mechanism is still able to identify salient sentences,
since similar idea in LexRank and TextRank methods achieves good performance
on MDS. Therefore, the attention distribution is now computed on all
the input sentences, by conducting the topic-sensitive PageRank algorithm
in Eq.~\ref{eq:close-form} and Eq.~\ref{eq:diffatt} on all the
original sentences.

However, a problem does occur because the amount of original sentences
is much larger than that of single document summarization task. Even
though the graph-based attention mechanism is still able to rank the
relevance and salience of original sentences, the attention distribution
will be too disperse and even. This results in that too many sentences
are considered to produce the context vector, making the context vector
contain too much information. We believe a more concentrated attention
distribution will be better. Therefore, when computing the attention
weights, only the top $K$ ranked sentences can have attention weights.
This can be easily realized by switching the rank scores of sentences
not in largest $K$ sentences to minimum value and re-normalizing
the attention weights. $K$ is a hyper-parameter.

\subsection{Model Tuning}

SinABS is trained on the single document summarization corpus - CNN/DailyMail.
Although both the CNN/DailyMail corpus and DUC datasets are news data,
the reference summaries of the datasets differ much. In order to better
adapt the SinABS model on the MDS task, we attempt to fine tune the
pre-trained SinABS model, although we have only a few reference summaries
for the MDS task. In our approach we tune the decoders of the model.
The parameters are the LSTM parameters of the word and sentence decoders,
and the weight vector $\mathbf{q}$ in the document set encoder. The
loss function and the optimization algorithm are the same with those
of the original SinABS model, and we use the cross-entropy loss and
the Adam \cite{kingma2014adam} algorithm to train the model. To prevent
overfitting the training is stopped when performance begins to decrease.

\section{Experiments}

\subsection{Dataset}

We conduct experiments on the DUC datasets which are widely used in
document summarization. We use the MDS tasks of DUC 2002 and 2004
as test sets, which contain 50 document sets and 59 document sets,
respectively. When evaluating on the DUC 2004 dataset, the DUC 2001-2003
and DUC 2005-2007 datasets are used for tuning the model, and DUC
2001, DUC 2003-2007 datasets are used when testing on the DUC 2002
dataset. The MDS tasks of DUC 2005-2007 are query focused summarization,
but we ignore the query since these datasets are only used for training.
There are on average 10 documents per set in DUC 2004 and 9.58 documents
per set in DUC 2002. For the datasets of DUC 2005-2007 we use only
the top 10 documents which are most similar to the topic of a document
set.

\subsection{Implementation}

We implement our approach based on the source code and pre-trained
model on the CNN/DailyMail corpus provided by \newcite{tan2017abstractive}.
We process the DUC datasets similar to \newcite{tan2017abstractive},
including tokenizing and lowercasing the text, replacing all digit
characters with the ``\#\textquotedblright{} symbol and label all
name entities with CoreNLP toolkit\footnote{http://stanfordnlp.github.io/CoreNLP/}.
The ``\#\textquotedblright{} symbols are mapped back to the original
digits after decoding according to the context. We also implement
our model in Theano\footnote{https://github.com/Theano/Theano} based
on the SinABS model. $K$ is set to 15 based on developing on the
training set.

\subsection{Evaluation Metric}

\textbf{~~~~ROUGE}: We use ROUGE-1.5.5 \cite{lin2003automatic}
toolkit and report the Rouge-1, Rouge-2 and Rouge-SU4 F1-scores, which
has been widely adopted by DUC and TAC for automatic summary quality
evaluation. It measured summary quality by counting overlapping units
such as the $n$-gram, word sequences and word pairs between the candidate
summary and the reference summary.

\textbf{Edit distance}: In order to test if our model is truly abstractive,
instead of simply copying relevant fragments verbatim from the input
documents, we compute the word edit-distance between each generated
sentence $s_{i}$ and the most similar original sentence of it, as
$ed_{i}$, and report the average $\textrm{ED}=\frac{1}{n}\sum_{i=1}^{n}ed_{i}$.

Considering the significant difference of length between sentences,
we also divide the word edit-distance for each generated sentence
by its word number $w_{i}$ as $\textrm{ED/}w=\frac{1}{n}\sum_{i=1}^{n}\nicefrac{ed_{i}}{w_{i}}$.

\subsection{Baselines}

To verify the effectiveness of our approach, we investigate various
strategies to adapt SinABS to MDS task for comparison. Since SinABS
takes one document as input but there are multiple input documents
in the MDS task, we explore four possible approaches to address this
(``ex.'' indicates extractive method and ``ab.'' indicates abstractive
method. SinABS is denoted as $\Delta$).

\textbf{Single-ab.}: One representative document of every document
set is selected as the input document to the SinABS model. This is
the most straightforward way to adapt single document summarization
model to the MDS task. The representative document is chosen by conducting
the PageRank \cite{page1999pagerank} algorithm on every document
set. This baseline is denoted as P.R.+$\Delta$.

\textbf{Single-ex.+Merge+Single-ab.}: Different from selecting one
representative document, we also investigate constructing a pseudo
document as the input to SinABS. We achieve this by first using extractive
single document summarization method to summarize every input document,
and then concatenate these summaries to form a new document. The motivation
of this strategy is to keep only the important content of original
documents, so that the input is both the key information and suitable
for SinABS to handle. The methods for extractive summarization are
Lead, LexRank, TextRank and Centroid. These four baselines are denoted
as Lead/Lex./Text./Cent.+$\Delta$ respectively.

\textbf{Single-ab.+Merge+Single-ab.}: Generate the abstractive summary
for every original document with SinABS. Then the abstractive summaries
are concatenated to form a pseudo document, as the input to SinABS
again. The difference from Single-ex.+Merge+Single-ab. is that no
extractive methods are required. This baseline his denoted as $\Delta$+$\Delta$.

\textbf{Single-ab.+Multi-ex.}: Generate the summary for every original
document, then summarize these summaries using some extractive MDS
method instead of SinABS to get the final summary. The extractive
MDS methods used are Lead, LexRank, TextRank, Centroid and Coverage.
Note that Coverage is specially designed for the MDS task, therefore
it is not used in Single-ex.+Merge+Single-ab. baselines. These five
baselines are denoted as $\Delta$+Lex./Text./Cent./Cov./Lead.

We introduce the extractive MDS methods used in previous baselines
as follows. These extractive methods themselves can also be the baselines
for comparison.

\textbf{Lead}: This baseline method takes the first sentences one
by one in single document or the first document in the document collection,
where documents in the collection are assumed to be ordered by name.

\textbf{Coverage}: It takes the first sentence one by one from the
first document to the last document in the document collection.

\textbf{LexRank}: LexRank \cite{erkan2004lexrank} computes sentence
importance based on the concept of eigenvector centrality in a graph
representation of sentences. In this model, a connectivity matrix
based on intra-sentence cosine similarity is used as the adjacency
matrix of the graph representation of sentences.

\textbf{TextRank}: TextRank \cite{mihalcea2004textrank} builds a
graph and adds each sentence as vertices, the overlap of two sentences
is treated as the relation that connects sentences. Then graph-based
ranking algorithm is applied until convergence. Sentences are sorted
based on their final score and a greedy algorithm is employed to impose
diversity penalty on each sentence and select summary sentences.

\textbf{Centroid}: In centroid-based summarization \cite{radev2000centroid}
method, a pseudo-sentence of the document called centroid is calculated.
The centroid consists of words with TF-IDF scores above a predefined
threshold. The score of each sentence is defined by summing the scores
based on different features including cosine similarity of sentences
with the centroid, position weight and cosine similarity with the
first sentence.

\begin{table}[ht]
\begin{tabular*}{1\columnwidth}{@{\extracolsep{\fill}}@{\extracolsep{\fill}}@{\extracolsep{\fill}}lccccc}
\toprule
\textbf{Method}  & \textbf{R-1}  & \textbf{R-2}  & \textbf{R-SU4}  & \textbf{ED}  & \textbf{ED/w}\tabularnewline
\midrule
P.R. +$\Delta$  & 28.3  & 4.83  & 8.8  & 24  & 0.88\tabularnewline
\midrule
Lead+$\Delta$  & 31.9  & 5.85  & 10.1  & 30  & 0.87\tabularnewline
Lex.\,+$\Delta$  & 31.0  & 5.52  & 9.8  & 25  & 0.87\tabularnewline
Text.+$\Delta$  & 32.3  & 5.68  & 10.4  & 34  & 0.89\tabularnewline
Cent.+$\Delta$  & 32.4  & 6.42  & 10.4  & 31  & 0.90\tabularnewline
\midrule
$\Delta$+Lead  & 31.5  & 5.34  & 9.9  & 27  & 0.87\tabularnewline
$\Delta$+Cov.  & 32.4  & 5.65  & 10.3  & 29  & 0.88\tabularnewline
$\Delta$+Lex.  & 32.7  & 5.80  & 10.5  & 20  & 0.96\tabularnewline
$\Delta$+Text.  & 32.6  & 5.96  & 10.4  & 32  & 0.79\tabularnewline
$\Delta$+Cent.  & 31.7  & 5.44  & 10.0  & 43  & 0.80\tabularnewline
\midrule
$\Delta$+$\Delta$  & 31.5  & 5.30  & 10.0  & 48  & 0.88\tabularnewline
\midrule
Our Model  & \textbf{34.0}  & \textbf{6.96}  & \textbf{11.4}  & 22  & 1.01\tabularnewline
\bottomrule
\end{tabular*}\caption{\label{tab:abbase2002}Comparison results with abstractive baselines
on the DUC 2002 test set.}
\end{table}

\subsection{Results}

\begin{table}[ht]
\begin{tabular*}{1\columnwidth}{@{\extracolsep{\fill}}@{\extracolsep{\fill}}@{\extracolsep{\fill}}lccccc}
\toprule
\textbf{Method}  & \textbf{R-1}  & \textbf{R-2}  & \textbf{R-SU4}  & \textbf{ED}  & \textbf{ED/w}\tabularnewline
\midrule
P.R. +$\Delta$  & 31.7  & 5.56  & 10.1  & 27  & 0.85\tabularnewline
\midrule
Lead+$\Delta$  & 31.8  & 5.74  & 10.0  & 28  & 0.83\tabularnewline
Lex.\,+$\Delta$  & 32.9  & 6.28  & 10.8  & 33  & 0.89\tabularnewline
Text.+$\Delta$  & 33.3  & 6.10  & 10.7  & 41  & 0.90\tabularnewline
Cent.+$\Delta$  & 34.4  & 6.68  & 11.1  & 44  & 0.93\tabularnewline
\midrule
$\Delta$+Lead  & 33.2  & 6.12  & 10.6  & 27  & 0.83\tabularnewline
$\Delta$+Cov.  & 34.4  & 6.84  & 11.2  & 27  & 0.84\tabularnewline
$\Delta$+Lex.  & 34.0  & 6.30  & 11.0  & 20  & 0.91\tabularnewline
$\Delta$+Text.  & 34.3  & 6.71  & 11.1  & 35  & 0.78\tabularnewline
$\Delta$+Cent.  & 32.8  & 5.77  & 10.3  & 44  & 0.80\tabularnewline
\midrule
$\Delta$+$\Delta$  & 31.3  & 4.70  & 9.6  & 52  & 0.88\tabularnewline
\midrule
Our Model  & \textbf{36.7}  & \textbf{7.83}  & \textbf{12.4}  & 22  & 1.10\tabularnewline
\bottomrule
\end{tabular*}\caption{\label{tab:abbase2004}Comparison results with abstractive baselines
on the DUC 2004 test set.}
\end{table}

The comparison results with abstractive baselines are presented in
Table~\ref{tab:abbase2002} and Table~\ref{tab:abbase2004}, respectively.
As seen from Table~\ref{tab:abbase2002} and Table~\ref{tab:abbase2004},
selecting one document as the representation of a document set (Single-ab.)
performs poorly. This indicates considering the information of all
documents is necessary for MDS task. Generally generating the abstractive
summary for every document first and then merging these summaries
with extractive MDS methods (i.e. Single-ab.+Multi-ex.) performs slightly
better than constructing pseudo single document by extractive summarization
methods (i.e. Single-ex.+Merge+Single-ab.). It may be explained that
Single-ab.+Multi-ex. keeps the integrity of a document, thus the SinABS
model will perform better. Similarly Single-ab.+Merge+Single-ab. does
not perform well because the constructed document is much different
from a real one. Our system achieves the best performance on both
datasets, since our model at the same time keeps the integrity of
all original documents and takes into consideration only the salient
sentences by ranking all original sentences in the attention mechanism.

The edit distance results verify that our method produces sentences
that are quite different from original sentences, indicating the property
of abstractive summarization.

\subsection{Model Validation}

\begin{table}
\begin{tabular*}{1\columnwidth}{@{\extracolsep{\fill}}@{\extracolsep{\fill}}@{\extracolsep{\fill}}lccc}
\toprule
\textbf{Method} & \textbf{Encoder} & \textbf{Attention} & \textbf{Tuning}\tabularnewline
\midrule
Model-1 & fixed & raw & no\tabularnewline
Model-2 & fixed & concentrated & no\tabularnewline
Model-3 & fixed & concentrated & yes\tabularnewline
Our Model & learned & concentrated & yes\tabularnewline
\bottomrule
\end{tabular*}

\caption{\label{tab:modelmeans} Details of model validation.}
\end{table}

We conduct ablation experiments to verify the effectiveness of our
model. Since we make three extensions to the SinABS model, namely
the learned weights in the document set encoder, the attention mechanism
and the tuning of the model. We validate their effect with three baseline
models, by each changes one of the three parts. The difference of
the three baselines are listed in Table~\ref{tab:modelmeans}. Model-1
is the simplest model without tuning, which uses a fixed weight vector
$\mathbf{w}=[\nicefrac{1}{m},\ldots,\nicefrac{1}{m}]$, and uses the
raw attention mechanism in \newcite{tan2017abstractive}. Model-2
verifies the effectiveness of making the attention distribution more
concentrated on the 15 most salient sentences. Model-3 verifies tuning
the decoder but not the document set encoder. Compared with Model-3, our
model further learns different weights for different documents in
the document encoder. Results are presented in Table~\ref{tab:model2002}
and Table~\ref{tab:model2004}. As seen from Table~\ref{tab:model2002}
and Table~\ref{tab:model2004}, all the three strategies considerably
improve the performance, validating how to better adapt single abstractive
summarization model to the MDS task.

\begin{table}
\begin{tabular*}{1\columnwidth}{@{\extracolsep{\fill}}@{\extracolsep{\fill}}@{\extracolsep{\fill}}lccccc}
\toprule
\textbf{Method} & \textbf{R-1} & \textbf{R-2} & \textbf{R-SU4} & \textbf{ED} & \textbf{ED/w}\tabularnewline
\midrule
Model-1 & 31.7 & 5.89 & 10.0 & 42 & 0.89\tabularnewline
Model-2 & 32.2 & 6.16 & 10.3 & 43 & 0.90\tabularnewline
Model-3 & 32.8 & 6.42 & 10.8 & 24 & 1.06\tabularnewline
Our Model & \textbf{34.0} & \textbf{6.96} & \textbf{11.4} & 22 & 1.01\tabularnewline
\bottomrule
\end{tabular*}\caption{\label{tab:model2002} Model validation results on DUC 2002.}
\end{table}

\begin{table}[ht]
\begin{tabular*}{1\columnwidth}{@{\extracolsep{\fill}}@{\extracolsep{\fill}}@{\extracolsep{\fill}}lccccc}
\toprule
\textbf{Method} & \textbf{R-1} & \textbf{R-2} & \textbf{R-SU4} & \textbf{ED} & \textbf{ED/w}\tabularnewline
\midrule
Model-1 & 33.9 & 6.64 & 11.0 & 45 & 0.90\tabularnewline
Model-2 & 34.1 & 7.10 & 11.2 & 49 & 0.91\tabularnewline
Model-3 & 34.9 & 7.52 & 11.8 & 21 & 1.06\tabularnewline
Our Model & \textbf{36.7} & \textbf{7.83} & \textbf{12.4} & 22 & 1.10\tabularnewline
\bottomrule
\end{tabular*}\caption{\label{tab:model2004}Model validation results on DUC 2004.}
\end{table}

\subsection{Human Evaluation}

We also conduct human evaluation to evaluate the linguistic quality
of the generated abstractive summaries, and compare with some significant
baselines. We randomly sample 10 document sets from the DUC 2002 dataset
and another 10 document sets from the DUC 2004 dataset for human evaluation.
Three volunteers who are fluent in English were asked to perform manual
ratings on three dimensions: Coherence, Non-Redundancy (N.R. for short)
and Readability. The ratings are in the format of 1-5 numerical scores
(not necessarily integral), with higher scores denote better quality.
The results are shown in Table~\ref{tab:manual2002}. It can be observed
that our system also outperforms other abstractive summarization approaches
in human evaluation, achieving good coherence and readability.

\begin{table}
\begin{tabular*}{1\columnwidth}{@{\extracolsep{\fill}}@{\extracolsep{\fill}}@{\extracolsep{\fill}}lccc}
\toprule
\textbf{Method} & \textbf{Coherence} & \textbf{N.R.} & \textbf{Readability}\tabularnewline
\midrule
Lead+$\Delta$ & 2.32 & 2.74 & 2.71\tabularnewline
Cent.+$\Delta$ & 2.63 & 2.84 & 3.29\tabularnewline
$\Delta$+Cov. & 2.30 & 3.53 & 2.92\tabularnewline
$\Delta$+Text. & 3.18 & 3.75 & 3.34\tabularnewline
$\Delta$+$\Delta$ & 2.23 & 2.57 & 2.57\tabularnewline
Our Model & \textbf{3.76} & \textbf{3.92} & \textbf{4.08}\tabularnewline
\bottomrule
\end{tabular*}\caption{\label{tab:manual2002}Human evaluation results on 20 samples from
the DUC 2002 and DUC 2004 datasets.}
\end{table}

\subsection{Case Study}

We show the abstractive summaries generated for an example from the
DUC 2004 test set in Figure~\ref{fig:Case-study}. It can be seen
that the abstractive summaries generally read well, and has the potential
to better convey the key information of original documents.

\begin{figure}
\noindent\fbox{\begin{minipage}[t]{1\columnwidth - 2\fboxsep - 2\fboxrule}%
\begin{scriptsize}

\textbf{\textit{Lead+}}$\Delta$:

politics , opposition leader hun sen and the prime minister were ousted
\textless{}eos\textgreater{} in the u.s. khmer rouge , the government
's prime minister 's ruling party has had a lengthy majority of its
leader in cambodia 's human rights record . \textless{}eos\textgreater{}
of the country 's opposition party leaders and opposition members
, the government have become prime minister \textless{}eos\textgreater{}
of parliament with its prime minister , the presidency of the khmer
rouge has been ruled out by the government 's leading opposition \textless{}eos\textgreater{}
two political parties previously clashed with the government 's top
two parties \textless{}eod\textgreater{}

\rule[0.5ex]{1\columnwidth}{1pt}

$\Delta$\textbf{\textit{+Text}}:

king hun sen on tuesday praised by cambodia 's top two political parties,
a coalition government led by prime minister \textless{}eos\textgreater{}
in a short letter sent to news agencies, the king said he had received
copies of fiscal and his cambodian people 's party in the government.
\textless{}eos\textgreater{} cambodia 's leading opposition party
ruled out sharing the top position in the presidency of parliament
with its opposition \textless{}eos\textgreater{} in talks between
the two party opposition bloc and the cambodian people 's party to
form a new government. \textless{}eod\textgreater{}

\rule[0.5ex]{1\columnwidth}{1pt}

\textbf{\textit{Our System}}:

opposition leader cambodian people 's party won the election. \textless{}eos\textgreater{}
in the u.s. , they were arrested in bangkok and charged with a lengthy
coup of human rights . \textless{}eos\textgreater{} leading opposition
party , the top position in parliament with its political rights ,
was arrested in bangkok , insisting it would lead to the presidency
of thailand 's leading government . \textless{}eos\textgreater{} prime
minister , political parties won a three - month agreement and agreed
to a coalition government . \textless{}eos\textgreater{} the government
would not end in a new coup vote and his arrest was rejected by the
parties of parliament . \textless{}eod\textgreater{}

\end{scriptsize}%
\end{minipage}}

\caption{\label{fig:Case-study}Example of generated abstractive summary by
our system. }
\end{figure}

\section{Conclusion and Future Work}

Abstractive Multi-Document Summarization (MDS) is still a challenging
and open problem. Although sequence-to-sequence models have achieved
great progress in single document summarization, the demands of large
amount of training data makes it hard to apply it to the MDS task.
In this paper, we address this problem from another direction, that
we investigate leveraging pre-trained successful single document summarization
model to the MDS task. We propose a framework to realize this goal
by adding a document set encoder into the hierarchical framework, and
we propose three strategies to further improve the model performance.
Experimental results demonstrate our approach is able to achieve promising
results on standard MDS datasets.

Our study is still primary effort towards abstractive MDS. Future
work we can do includes alleviating the requirement of a good pre-trained
abstractive summarization model, designing better attention mechanism
for MDS, and investigating our approach based on other model architectures.

 \bibliographystyle{acl_natbib}

\end{document}